\documentclass[%
 reprint,
nofootinbib,
 amsmath,amssymb,https://es.overleaf.com/1666689368hgdkfdcbjvmv
 aps,
]{revtex4-2}

\usepackage{graphicx}
\usepackage{dcolumn}
\usepackage{bm}
\usepackage[utf8]{inputenc} 
\usepackage[T1]{fontenc}    
\usepackage{hyperref}       
\usepackage{url}            
\usepackage{booktabs}       
\usepackage{amsfonts}       
\usepackage{nicefrac}       
\usepackage{microtype}      

\usepackage{xspace}
\usepackage{stmaryrd}

\usepackage{times}
\usepackage{soul}
\usepackage[small]{caption}
\usepackage{amsmath}
\usepackage{booktabs}
\usepackage{algorithm}
\usepackage{algorithmic}
\usepackage{xspace}
\usepackage{multirow,tabularx}
\usepackage{subcaption}
\usepackage{tikz}
\usepackage{pgfplots}
\pgfplotsset{compat=1.10}
\usetikzlibrary{shapes.geometric,arrows,fit,matrix,positioning}
\tikzset
{
    treenode/.style = {rectangle, draw=black, align=center, minimum size=1cm},
}

\newcommand{\start}{\textsc{start}}
\newcommand{\bool}{\textsc{bool}}
\newcommand{\atom}{\textsc{atom}}
\newcommand{\oxor}{\oplus}
\newcommand{\oand}{\wedge}
\newcommand{\oor}{\vee}
\newcommand{\grambool}{\ensuremath{{\sf P}}\xspace}
\newcommand{\gramboolxor}{\ensuremath{{\sf P}^\oxor}\xspace}
\newcommand{\controla}{$\con^1$\xspace}
\newcommand{\controlb}{$\con^2_c$\xspace}
\newcommand{\controlc}{$\con^3_c$\xspace}
\newcommand{\controld}{$\con^4_c$\xspace}
\newcommand{\targeta} {$\con^1$\xspace}
\newcommand{\targetb} {$\con^2_t$\xspace}
\newcommand{\targetc} {$\con^3_t$\xspace}
\newcommand{\targetd} {$\con^4_t$\xspace}
\newcommand{\testa}   {$\con^5$\xspace}
\newcommand{\testb}   {$\con^6$\xspace}

\begin{document}

\preprint{APS/123-QED}

\title{Towards a more flexible Language of Thought: \\ Bayesian grammar updates after each concept exposure}

\author{Pablo Tano$^1$}
\author{Sergio Romano$^{2,3}$ }
\author{Mariano Sigman$^4$}
\author{Alejo Salles$^1$}
\author{Santiago Figueira$^{2,3}$}
\affiliation{
$^1$CONICET-Universidad de Buenos Aires. Instituto de Cálculo. Argentina \\
$^2$CONICET-Universidad de Buenos Aires. Instituto de Ciencias de la Computación. Argentina. \\
$^3$Universidad de Buenos Aires. Facultad de Ciencias Exactas y Naturales. Argentina \\
$^4$Universidad Torcuato di Tella. Argentina.
}
 

\begin{abstract}
Recent approaches to human concept learning have successfully combined the power of symbolic, infinitely productive rule systems and statistical learning to explain our ability to learn new concepts from  just a few examples. The aim of most of these studies is to reveal the underlying language structuring these representations and providing a general substrate for thought. However, describing a model of thought that is fixed once trained is against the extensive literature that shows how experience shapes concept learning. Here, we ask about the plasticity of these symbolic descriptive languages. We perform a concept learning experiment that demonstrates that humans can change very rapidly the repertoire of symbols they use to identify concepts, by compiling expressions which are frequently used into new symbols of the language. The pattern of concept learning times is accurately described by a Bayesian agent that rationally updates the probability of compiling a new expression according to how useful it has been to compress concepts so far. By portraying the Language of Thought as a flexible system of rules, we also highlight the difficulties to pin it down empirically.
\end{abstract}

\maketitle


\section{Introduction}

How can children acquire a vast universe of concepts with seemingly very little exposure? One possible solution to this conundrum, known as the Plato Problem~\cite{chomsky1986knowledge,chomsky2006cognitive}, builds on the human capacity to describe concepts --and more generally of all elements of thought-- through the use of a symbolic and combinatorial mental language~\cite{newell1980physical}, referred as {\em language of thought} (LoT)~\cite{fodor1975language}.

Combinatorial languages can describe a vast set of concepts from a small set of primitives. This can be understood in a relatively simple example in the domain of shapes. A combinatorial and symbolic language similar to Logo~\cite{abelson1974logo} can combine operations such as ``move", ``pen up", ``pen down" or ``rotate" to generate an infinite set of expressions (or programs) which, when evaluated, can convey all sort of shapes.

A language describing concepts (like shapes) also provides a natural notion of their complexity \cite{kolmogorov1968three}. A concept is simple, relative to that language, when it can be described by a short program. On the contrary, it is complex when all its descriptions require a long sequence of instructions. For example, in the case of the Logo language, a square can simply be instructed as a loop of four displacements followed by rotations of 90 degrees. In this language, the icon of a face will be implemented by a significant lengthier program and hence will be more complex.  However, this concept would be simpler when described in a language in which the icon of a face (or the symbols for nose, mouth, etc.) are available as primitives in the language.

In the domain of Boolean concepts, a wide range of logical varieties of concepts was studied in~\cite{feldman2003simplicity}, revealing a surprisingly simple `law': the subjective difficulty of a Boolean concept for a human learner is directly proportional to the length of the shortest compatible program in the language of propositional logic (i.e.\ Boolean variables combined with the operators \textit{and}, \textit{or} and \textit{not}). This result may suggest that human LoT is equipped with rules and symbols similar to those found in propositional logic. Indeed, the correlation between the subjective difficulty of concepts and their complexity has been used as a general vehicle to study human LoT in various domains ~\cite{piantadosi2016logical,leeuwenberg1971perceptual,amalric2017language,romano2018,lupyan2007language}. Although often implicit, the general strategy is to (\textit{1)} assume a language; (\textit{2)} find the shortest compatible program for some concepts in that language; (\textit{3)} compare the length of these programs with the subjective difficulty of the concepts; and finally (\textit{4)} repeat this process for various languages within a universe of possible candidates and choose the language that gives the best match in \textit{(3)}. As mentioned before, the length of the program depends on the primitives of the language in which this program is written, so different languages make different predictions.

A natural question, however, is whether the primitives of a LoT are universal --both across different individuals and also throughout development-- or if instead the semantic repertoire of a language is dynamic and shaped by experience. Indeed, it is likely that our ability to automatically represent Boolean concepts in a succinct manner is not due to an innate efficient propositional language in our mind. Instead, we propose that this ability arises as a byproduct of our brain rapidly learning efficient representations for the concepts we usually  encounter in everyday life. Our research question is: how rapidly can we adapt our learning mechanisms when we encounter a new domain in which our a priori representations are no longer efficient? We examine the hypothesis that humans have the ability to rapidly recombine propositions in their LoT, adding new primitives to their language. In other words, that learning leads to a process of compiling routines into functions within the LoT.

In the example of the Logo language one can imagine that if productions which draw squares are very frequent, it would be efficient to devote a new symbol to this production. The new symbol `square' is a hierarchical `second order' construction of the `first order' primitives of the language. It has a cost (of increasing the lexicon of the language) but in the new language, drawing a square can be instantiated with a very short program (namely, `square') and hence uses less memory. Indeed, a higher level language allows us to reach a higher level of abstraction by freeing memory and processing power, thus making more complex thoughts thinkable~\cite{minsky1967computation,murphy1988comprehending}.

Most work in the LoT literature, while naturally including a learning mechanism, tends to approach the LoT as a stable system to be unearthed by experimenters, who try different candidate templates and select the one which best fits the data after training ~\cite{goodman2008rational,kemp2012exploring,piantadosi2016logical}. Still, how different tracks of experience can shape acquisition differently and can constantly change the repertoire of a LoT after each exposure remains to be discovered.

Here, we perform a Boolean concept learning experiment to show that humans can change very rapidly --in the course of an experiment-- the repertoire of symbols they use to identify concepts. We also provide a dynamic model that is flexible enough to update its underlying language after each concept exposure. 

In our experiment, participants are divided in two groups, in such a way that each group is presented with a different sequence of concepts. One of the two groups is presented with concepts that are succinctly described only if the logical operator `exclusive or' (xor, notated~$\oxor$) is used, which we presume does not form part of the natural repertoire of LoT in this specific domain~\cite{piantadosi2016logical}. However, these concepts can also be described with a sensibly lengthy combination of primitives excluding $\oxor$. We show how the exposure to this set of concepts `compiles' the $\oxor$ operator in a way that, after exposure, subjective difficulty is described by an extended language in which $\oxor$ has been incorporated to the set of primitives. Furthermore, we show that the subjective difficulty of concepts throughout the task is consistent with that of a Bayesian agent that rationally updates the probability of compiling $\oxor$ according to how useful it has been to compress concepts so far.

\newcommand{\sem}[1]{\llbracket #1\rrbracket}
\newcommand{\vars}{{\sf Vars}}
\newcommand{\con}{{\mathcal C}}
\newcommand{\mdl}[1]{\textsc{mdl}_{#1}}

\section{The logical setting}

We consider two propositional logics, both containing only four  propositional variables $\vars=\{x_1, x_2, x_3, x_4\}$. \grambool is defined over the signature $\oand$, $\oor$ and $\lnot$, and \gramboolxor is defined over the signature $\oand$, $\oor$, $\lnot$ and $\oxor$. As one can see from the grammars defined in Fig.~\ref{PCFG}, the only difference between \grambool and \gramboolxor is that the latter has an additional operator $\oxor$.  
 \begin{figure}[h!]
\centering
\small\vspace{-.3cm}
\begin{tabular}{ccc}
\begin{minipage}[h]{0,2\textwidth}
\begin{eqnarray*}
\start &\to&\bool\\
\bool &\to&(\bool \oand \bool) \\
\bool &\to&(\bool \oor \bool) \\
\bool &\to&\atom
\end{eqnarray*}
\end{minipage}
&
\ \quad
&
\begin{minipage}[h]{0,2\textwidth}
\ \quad For $i=1,2,3,4$
\begin{eqnarray*}
\atom &\to& x_i \\
\atom &\to&\lnot x_i 
\end{eqnarray*}
\end{minipage}
\end{tabular}
      \caption{The context free grammar for language \grambool.  Language \gramboolxor has an extra rule: $\bool\  \to\ (\bool \oxor \bool)$}
      \label{PCFG}
   \end{figure}

The semantics of $\oand$, $\oor$ and $\lnot$ are standard: conjunction, disjunction and negation, respectively. We let $\oxor$ denote the exclusive disjunction. As usual, $v\models \varphi$, represents that the formula $\varphi$ is true for the valuation $v:\vars\to\{0,1\}$ and we denote the {\em semantics} of $\varphi$ by $\sem{\varphi}=\{v\colon v\models\varphi\}$. A {\em concept} $\con$ is a set of valuations $\vars\to\{0,1\}$. The complement of $\con$ is denoted $\overline \con$ and is defined as $\overline \con=\{0,1\}^\vars\setminus \con$. Observe that $\#\con+\#\overline \con=16$.
We say that a formula $\varphi$ is {\em compatible} with concept $\con$ if $\sem{\varphi}=\con$. We regard logics as languages for describing concepts. Any concept $\con$ has infinitely many descriptions, namely, all formulas $\varphi$ such that $\sem{\varphi}=\con$. 

\begin{figure}[t!]
 \vspace{-0.5cm}
  \centering
  \includegraphics[scale=.335]{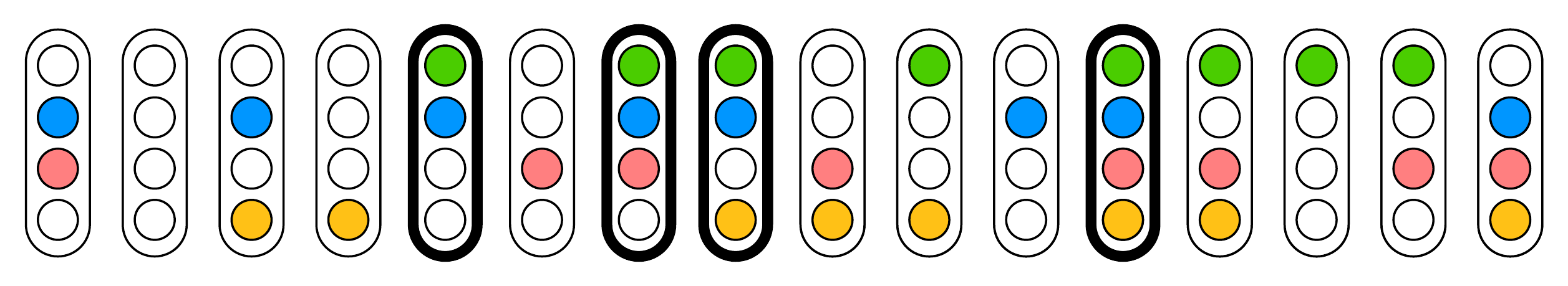}
  \caption{Example of a concept $\con$, as shown in the experiment. Variables in $\vars=\{x_1,x_2,x_3,x_4\}$ correspond to the presence of a  color light in the object ($x_{1}=$ green, $x_2=$ blue, $x_3=$ red, $x_4=$ orange). Items (valuations) belonging to $\con$ are highlighted with bold border. $\con$ may be described by $x_1\oand x_2$. }
  \label{semaforos}
\end{figure}

\paragraph*{Example.}
In Fig.~\ref{semaforos} we depict a concept $\con$ (variables are represented by colors) such that $\#\con=4$. One can see that the formula $x_3$ is not compatible with $\con$ but $x_1 \oand x_2$, or $x_1 \oand x_2 \oand (x_3 \oor \lnot x_3)$, are compatible with $\con$. $\overline \con$ may be described by $\lnot x_1\oor \lnot x_2$.

We will often identify concepts with any formula compatible with it, so we will talk of ``concept $\varphi$'' to refer to ``concept $\sem{\varphi}$''. However, it should be noted that a concept is a semantic object that has many descriptions in the logical language.

Our interest is in the size of the shortest descriptions of a given concept, that is, in the minimum description length (MDL) of concepts.

The {\em size} of a formula $\varphi$ is denoted $|\varphi|$ and it is defined as the number of operators plus the number of atoms (i.e.\ possibly negated propositional symbols), that is: $|x_i|=|\lnot x_i|=1$ for $i=1\dots 4$ and $|\varphi_1*\varphi_2|=|\varphi_1|+|\varphi_2|+1$ for $*\in\{\oand,\oor,\oxor\}$. For ${\sf L} \in\{\grambool,\gramboolxor\}$ and a concept $\con$ we define the {\em minimum description length of $\con$ with respect to $\mathcal {\sf L}$} as
$$
\mdl{\sf L}(\con)=\min\{|\varphi|\colon \varphi\in\mathcal L, \sem{\varphi}=\con\}.
$$
Since $\grambool$ is a sublanguage of $\gramboolxor$, we have $\mdl{\gramboolxor}(\con)\leq\mdl{\grambool}(\con)$ for any concept $\con$.
   
\paragraph*{Example.}The concept $
\con=\{v\colon v(x_1)+v(x_2)=1\},
$
which expresses that $x_1$ is true or $x_2$ is true but not both can be described in \gramboolxor as $\varphi=x_{1} \oxor x_{2}$, of length 3. In fact, one can check that this is the shortest formula compatible with $\con$, and so $\mdl{\gramboolxor}(\con)=3$. If we now switch to \grambool, we can no longer describe $\con$ as $x_{1} \oxor x_{2}$, since $\oxor$ is not part of its signature. However, in \grambool, the concept $\con$ may be described by  formula $\psi=(x_{1} \oand \neg x_{2}) \oor (x_{2} \oand \neg x_{1})$, of size 7. Since this formula has minimal size, we have that $\mdl{\grambool}(\con)=7$.

\section{Experiment}

\renewcommand*{\arraystretch}{1.2}
   
\begin{table*}[!ht]
\vspace{-0.5cm}
\centering

\begin{tabular}{|c|l l | c | c |l l | c | c|}
\hline
                                   &
\multicolumn{2}{c|}{\textbf{Target group}}       &
\textbf{$\mdl{\gramboolxor}(\con)$} &
\textbf{$\mdl{\grambool}(\con)$} &
\multicolumn{2}{c|}{\textbf{Control group}}  &
\textbf{$\mdl{\gramboolxor}(\con)$} &
\textbf{$\mdl{\grambool}(\con)$} 
 \\ \hline
\multirow{4}{*}{\parbox[t]{2mm}{\rotatebox[origin=c]{90}{\bf Training}}} & \targeta & $x_{i}$             & 1 & 1                                                                  & \multicolumn{4}{c|}{$\longleftarrow$Idem}                                                            \\ \cline{2-9}
                                   &\targetb& $x_{i} \oxor x_{j}$       & 3 & 7 &\controlb& $x_{i} \oor x_{j}$  & 3 & 3                                                                  \\ \cline{2-9}
                                  &\targetc & $x_{i} \oxor x_{j} \oxor x_{k}$ & 5 & 19 &\controlc& $x_{i} \oor (x_{j} \oand x_{k})$  & 5 & 5 \\ \cline{2-9}
                                  &\targetd& $x_{k} \oxor x_{l}$       & 3 & 7&\controld&$x_{k} \oor x_{l}$  & 3 & 3                                                                  \\ \hline
\multirow{2}{*}{\parbox[t]{2mm}{\rotatebox[origin=c]{90}{\bf Test}}}  &\testa& $x_{i} \oand (x_{j} \oxor x_{k})$ & 5 & 9                 & \multicolumn{4}{c|}{$\longleftarrow$Idem}                                                                                    \\ \cline{2-9}
                               &\testb& $x_{i} \oand (x_{j} \oor x_{k})$  & 5 & 5         & \multicolumn{4}{c|}{$\longleftarrow$Idem}                                                                                            \\ \hline
\end{tabular}

\caption{Sequence of concepts presented in the experiment: \targeta, \targetb, \targetc, \targetd, \testa, \testb for target group and  \controla, \controlb, \controlc, \controld, \testa, \testb for control group. Each concept $\con$ is represented by a minimal formula $\varphi$ such that $\sem{\varphi}=\con$. 
}
\label{conceptos}
\vspace{-0.4cm}
\end{table*}

55 participants participated in the experiment over the world wide web using the Amazon Mechanical Turk crowd sourcing platform. All were US residents over the age of 18 and had more than 95\% of past tasks successfully approved by other requesters. 44 participants completed the experiment through all the stages and declared not cheating (using pen, screenshots or a similar method to copy the answers) at the end of the experiment. Only data from these participants were used in the analyses reported below.\footnote{The learning times of all participants can be found in https://figshare.com/s/04d338adbbc4b1e83bf0.}

Participants were divided randomly into a control group ($N=21$) and a target group ($N=23$). Both groups were presented with different sequences of six concepts. For each concept, there was a learning phase, a testing phase and a feedback phase. The average time spent in each concept was 167$\pm$20 s.e.m.\ seconds, and the average duration of the task was 21$\pm$4 s.e.m.\ minutes. After moving through the learning, testing and feedback phase of each of the six concepts, participants were asked if they used a pen or recorded the screen information in any way. They were also told that the answer to this question will not affect their payment, but that it was crucial for the experimenters to know.

During the learning phase, all 16 items were presented in the screen (in random order), and items belonging to the concept were identified with bold boundaries, as shown in  Fig.~\ref{semaforos}.
Participants were told that only the items with bold boundaries were `blickets' (or `tufas', etc.: we used different words for each concept in the sequence), and asked them to try to identify what a blicket was. During the testing phase, the 16 items were shuffled in the screen, and participants were asked to click on items that were blickets. If they made mistakes after submitting their answer, they were directed to the feedback phase, in which items that were incorrectly classified were indicated with a red cross. After having studied the feedback, participants were redirected to the testing screen, where items were reshuffled. When every item was correctly classified, participants were asked to give a verbal description of the concept and then continued on to the following concept after a resting period. We characterize the subjective difficulty of each concept as the time the participant spent in learning, testing and feedback phases for that concept (excluding the time spent in the verbal description).

Both groups (target and control), were exposed to 6 concepts. The second, third and fourth concepts are {\em training} concepts, and were different between both groups. The last two concepts are the {\em test} concepts, and were the same for both groups. The first concept was the trivial concept $x_i$ for both groups, which was aimed to get participants started in the task. Importantly, variables (i.e. color lights inside objects in Fig.~\ref{semaforos}) were randomized for every concept, so paying selective attention to a specific variable across subsequent concepts was not beneficial for learning the concept sequence.

As shown in Table~\ref{conceptos}, we presented the target group with training concepts which are succinctly described when $\oxor$ is part of the language, but necessarily described with lengthier formulas if $\oxor$ is absent. More technically, concepts for which $\mdl{\gramboolxor}$ is much smaller than $\mdl{\grambool}$.

Participants in the control group, on the other hand, experienced a sequence of concepts that could be easily described using the language given by \grambool. After these training concepts, both groups were presented with the same pair of test concepts: one which could be only succinctly described in \gramboolxor, and one for which the MDL did not depend on the underlying language \gramboolxor or \grambool. We compared learning times between the two groups for these last two concepts.

As shown in Table~\ref{conceptos}, training concepts for the target (xor) group were: $x_{i}$, $x_{i} \oxor x_{j}$, $x_{i} \oxor x_{j} \oxor x_{k}$, and $x_{k} \oxor x_{l}$, called \targeta, \targetb, \targetc and \targetd respectively. Training concepts for the control group were: $x_{i}$, $x_{i} \oor x_{j}$, $x_{i} \oor (x_{j} \oand x_{k})$, and $x_{k} \oor x_{l}$ called \controla, \controlb, \controlc and \controld respectively. We use the indexes \textit{i, j, k, l} instead of numbers because variables were randomized in each trial. $x_{i}$ could stand for $x_1$, $x_2$, $x_3$ or $x_4$, that is, for any of the four colors. After these four concepts, both groups were presented with the same test concepts: $x_{i} \oand (x_{j} \oxor x_{k})$, and $x_{i} \oand (x_{j} \oor x_{k})$, called \testa and \testb respectively.

Choosing which concepts to show the target group in order for them to `learn' the $\oxor$ operator is critical in our experiment. Crucially, the learner must have an option between two alternatives that describe the concept: one that is succinct but uses $\oxor$, or necessarily a much longer one in the absence of $\oxor$. In other words, these concepts must be compatible with short logical formulas if and only if we take \gramboolxor as the language of description. To ensure that this was the case, we enumerated, for each concept, all formulas compatible with it and produced by the \grambool and \gramboolxor grammars up to length 19. For all training concepts of the target group, the shortest compatible formula without $\oxor$ is much longer than the shortest compatible formula with $\oxor$. This is shown in Table\ref{conceptos}.

\section{Model-Free Results}

We measure the subjective difficulty of a given concept as the total time needed by the participant to successfully encode the concept, which indicates that they can reliably express which exemplars belong to the concept and which do not.

\begin{figure}[t!]
      \includegraphics[scale=0.25]{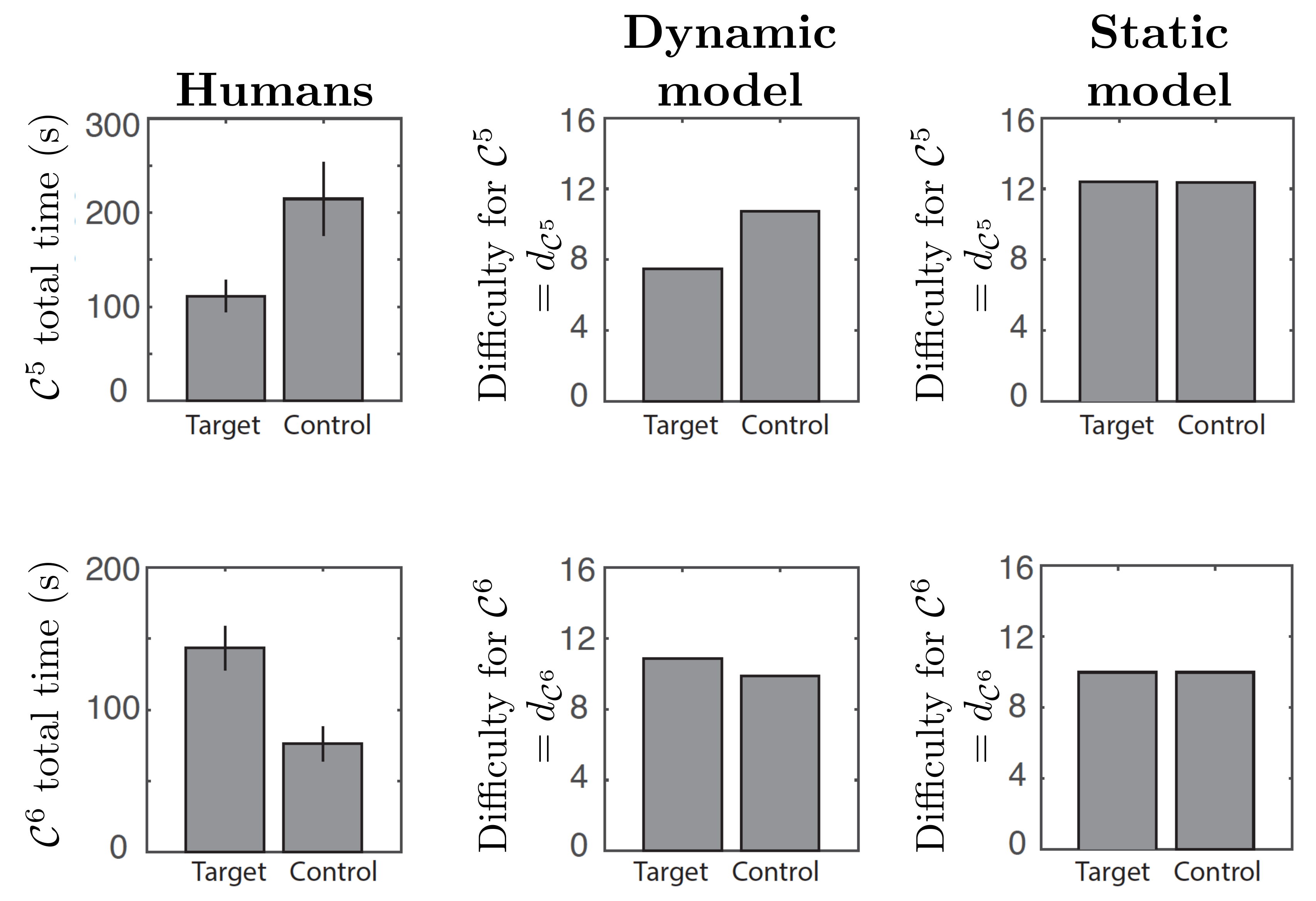}
      \centering
      \caption{Concept learning time (left) and difficulty predicted (middle, right) for the two test concepts (\testa and \testb). Error bars are s.e.m.\ across subjects.}
      \label{model free}
   \end{figure}

Participants from the target group spent almost half the time than participants from the control group in \testa, which could be succinctly described only in \gramboolxor ($111\pm16$ s.e.m.\ seconds versus $214\pm37$ s.e.m.\ seconds, a two-sample t-test reveals $t_{42}=2.6$, $P<0.01$), as shown in Fig.~\ref{model free} (left). We also found that the control group learned much faster \testb ($143\pm14$ s.e.m.\ seconds for the target group versus $76\pm10$ s.e.m.\ seconds for the control group, $t_{42}=3.5$, $P<0.01$). A mixed ANOVA with \testa-\testb as within subject factor and target-control groups as between subject factor reveals a strong interaction between group and \testa-\testb ($F=15.3$, $P<0.001$), indicating that the differences in learning times for \testa and \testb were very different between the two groups.

The target group encoded \testa more efficiently than the control group. We propose that the control group expected to find in \testa and \testb structures that could be easily built in \grambool. The target group, on the other hand, became biased towards the $\oxor$ structure, and they expected to find it in \testa and \testb. This caused \testa to be encoded more rapidly by the target group and \testb more rapidly by the control group. Assuming that the subjective difficulty of learning a concept is proportional to the complexity of its internal representation, we conclude that after exposure to the training concepts, participants in the target group represented the $\oxor$ more efficiently than the control group, and expected to find this structure in \testa and \testb.

\section{Model}

When presented with a concept (e.g.~Fig.~\ref{semaforos}), our model generates logical formulas and evaluate them to true or false for that concept, keeping the formula only if it is true. To generate formulas, the model uses a symbolic language in which each rule (symbols and operators) is associated with a probability of being used. The probability of generating a formula is proportional to the product of the probabilities of the rules required for building it, and therefore it decreases exponentially with its length. Furthermore, if one of the rules has a very low probability of being used, formulas that require it will also have very low probability.

The \textit{Static model} maintains the rules' probabilities fixed throughout the concept sequence (the 6 concepts in Table~\ref{conceptos}). The \textit{Dynamic model} updates the probabilities after each concept, in order to minimize the expected description length of future concepts, assuming they have similar structure to the concepts learnt so far. We include in this model the $\oxor$ rule a priori in the language, but with vanishing probability of being used. Changes in this probability can be analogously interpreted as the probability that a rational agent without the compiled symbol a priori decides to add the compiled expression as a new primitive into her language. 

\subsection{Static Model}

\newcommand{\form}{\varphi}
Under the LoT assumption,  given a concept $\con$ (e.g.\ Fig.~\ref{semaforos}), the probability that an agent uses formula $\form$ to explain this concept is defined by Bayes theorem: 
$$
P(\form\mid \con) \propto P(\con\mid \form)P(\form).
\label{Bayes fix}
$$
%
The likelihood $P(\con\mid \form)$ of a logical statement $\form$ can be simply defined as 1 if $\sem{\form}=\con$ and 0 otherwise. In other words, for any given concept, only explanations that describe this concept are considered as possible explanations. The likelihood term has been defined more flexibly in the literature~\cite{goodman2008rational,piantadosi2016logical}, allowing for mislabeled elements. We keep this simpler definition in order to reduce the number of free parameters of the model, as we do not intend to account for mislabeling errors in our experiment.

The prior $P(\form)$ is defined by augmenting the context-free grammars shown in Fig.~\ref{PCFG} into a probabilistic context-free grammars (PCFG). In the PCFG, each rule has associated a parameter indicating the probability of using that rule. A PCFG can be used to produce logical statements similar to a CFG. Each non-terminal remaining in the statement is expanded using a rule of the PCFG with probability proportional to that rule's associated parameter, until no non-terminals remain in the statement. 

We assume that the probability that a subject uses formula $\form$ to explain concept $\con$ is proportional to the posterior $P(\form \mid \con)$, and the subjective difficulty $d_\con$ of a concept $\con$ to a participant is proportional to the length of the formula that the participant is using to explain that concept. However, there is no way to know directly which internal formula $\form$ the participant is using (and therefore we do not know $|\form|$). Hence, the most parsimonious approach is to consider the entire posterior distribution $\textbf{P}(\form \mid \con)$ over possible formulas.\footnote{This is equivalent to the Sampling Hypothesis described in~\cite{denison2013rational}, by which participants represent distributions through samples. Similar results are obtained if each participant carries entire probability distributions.}

Given a concept $\con$, the expected length $E_\con$ of the formulas used by the participant is simply
 \begin{equation}
E_\con=\sum_{\sem{\form}=\con} |\form| \ P(\form \mid \con),
 \label{expected length}
 \end{equation}
where the sum is over all formulas $\varphi$ compatible with $\con$. We  define the difficulty $d_\con$ of a concept experienced by the participant  as $$d_\con \propto E_\con + \alpha N_\con,$$
%
%
where we added a term that accounts for the cardinality of the concept: $N_\con$ is the cardinality of the concept or its complement, the one being smaller, i.e.\ $N_\con=\min \{ \#\con, \#\overline \con\}$ (e.g.\ $N_\con=4$ for the concept $\con$ of  Fig.~\ref{semaforos}), and $\alpha$ is a free parameter fitted globally for all concepts and participants to its maximum likelihood value of 0.9. In this way, we remove the asymmetry between positive and negative examples, while accounting for the toil taken by considering a larger number of items simultaneously.

In practice, to approximate $E_\con$ for each concept $\con$, we calculated the posterior probability $P(\form\mid \con)$ of all compatible formulas $\form$s up to size 19 with $P(\form\mid \con)$ and then use ~\eqref{expected length}. Since the space of all possible $\form$s grows exponentially with $|\form|$, normative procedures for estimating $P(\form\mid \con)$ in this space involve stochastic search algorithms. However, in our case, we were able to exhaustively enumerate and calculate the posterior probability of \textit{all} formulas generated by the PCFG up to a sufficiently high size $M$ such that all formulas with $|\form|>M$ have vanishing probabilities when compared to shorter compatible formulas for the current concept (because the prior $P(\form)$ decreases exponentially with the size of the formula).

\subsection{Dynamic Model}

Up to this point, we assumed that, given a concept $\con$, the posterior distribution over formulas $P(\form\mid \con)$ was independent of the other concepts presented in the sequence. However, if the LoT (i.e. the PCFG) updates with experience, the prior $P(\form)$ in $P(\form\mid \con)$ will change, and so will $E_\con$ in \eqref{expected length} and finally the subjective difficulty $d_\con$. Therefore, $d_\con$ will depend on the sequence of concepts that were previously presented to the participant.

In other words, since now $P(\form)$ depends on the sequence of concepts experienced by the participant, instead of $P(\form\mid \con)$, we have $$P(\form\mid \con^{t},\dots,\con^{1}) \propto P(\con^{t}\mid \form)P(\form\mid \con^{1},\dots,\con^{t-1})$$, 
%
where $\con^{t}$ is the concept presented at trial~$t$, and $P(\form\mid \con^{1},\dots,\con^{t-1})$ depends on the state of the PCFG at trial $t$, which in turn depends on how the PCFG gets updated from trial to trial.

Intuitively, the update process increases the probability of using a certain rule in the PCFG accordingly to how useful this rule was to compress compatible formulas for the concepts previously learned in the same domain. Specifically, we model the update process in a normative manner: the probability of using a rule of the PCFG at trial $t$ is equal to the Bayesian posterior probability that this rule will enable the learner to find compressed explanations at trial $t$, according to how useful it was to compress explanations in trials $1,\dots,t-1$.

To formalize the update of the PCFG, we define $P(\form)$ similarly to~\cite{goodman2008rational}. Specifically, the prior probability of a logical statement at trial $t$ in the concept sequence uses a single Dirichlet-multinomial for the set of rule expansions. The Dirichlet is parameterized by a set of positive real numbers $D_{i}^{t}$, one for each rule $i$ in the PCFG, which in turn determine the probability of using rule $i$ at trial $t$: a higher $D_{i}$ indicates a higher probability of using rule $i$.

The prior is specified by the set Dirichlet parameters $\textbf{D}^{0}$ with which we start the experiment ($\textbf{D}^{0}$ represents a vector containing the prior parameters of all rules in the grammar at trial 0). In our experiment, we set the prior Dirichlet parameters of all rules equal to 1, and the parameter of the rule that expands the target operator to a value several orders of magnitude smaller ($\approx 10^{-4}$). This means that the target operator was practically absent at the beginning of the experiment, but it was technically possible to `learn it' by increasing its probability as the experiment developed.

Under the Dirichlet model, the prior $P(\form\mid \con^{1},\dots,\con^{t-1})$ can be rewritten using the Dirichlet parameters as $P(\form\mid\textbf{D}^{t})$. Therefore, to know how $P(\form\mid \con)$ updates from trial to trial, we only need to know how $\textbf{D}$ updates from trial to trial.

The Dirichlet parameter of rule $i$ at trial $t+1$ is equal to its parameter at trial $t$ plus the amount of times the production $i$ was used in generating all formulas compatible with the concept at trial $t$ (we note $M_{i}(\form)$ as the number of times that rule $i$ is used in generating formula $\form$), weighted by each formula's posterior probability at trial $t$:
 \begin{equation}
 D_{i}^{t+1}=D_{i}^{t}+\sum_{\sem{\form}=\con^t} P(\form\mid \textbf{D}^{t}) \ M_{i}(\form).
 \label{Dirichlet}
 \end{equation}
This Bayesian learning mechanism increases the probability of using rules that allow concepts to be succinctly described. This happens because these formulas have higher probability $P(\form\mid \textbf{D})$ than longer formulas, so the Dirichlet parameters of the rules that build these formulas increase more strongly than those of the rules that build longer formulas.   

\section{Results}

\begin{figure}
      \centering
      \includegraphics[scale=.30]{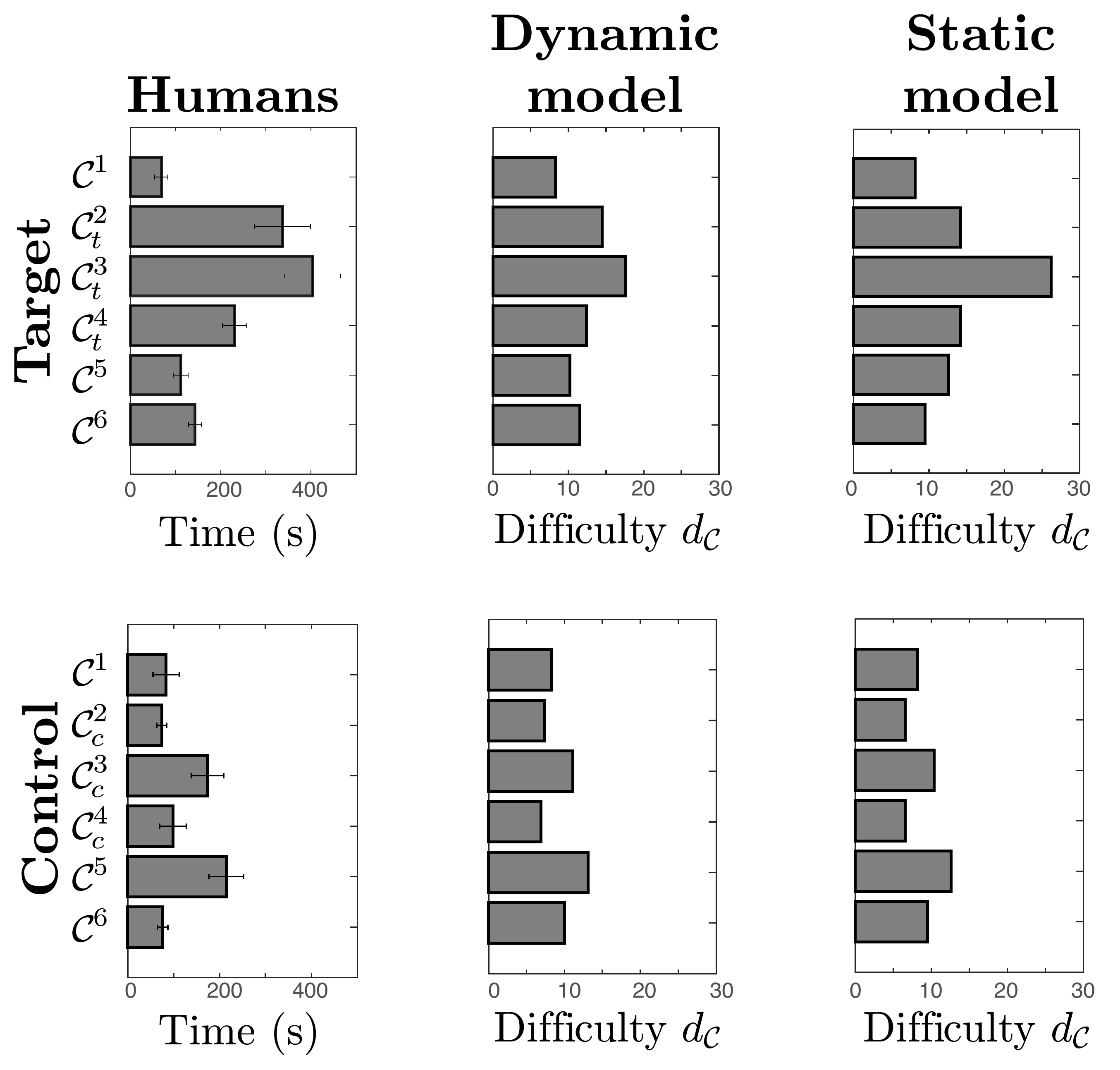}
      \caption{Learning times and model predictions for target and control groups (see Table~\ref{conceptos} for concept details). The predicted difficulties of each model were calculated using $d_\con$. Error bars are s.e.m.}
      \label{results}
\end{figure}

The Bayesian agent that minimizes the expected complexity of future concepts by optimally adapting its LoT to the inferred structure of the task accurately captures the dynamics of human learning across concepts. If we did not allow the model to update the probability of the operators after each concept, and particularly the compiled operator $\oxor$, the control group and the target group would be indistinguishable to the model as it would predict equal average formula length for both groups (see Fig.~\ref{model free}, {\em Static Model}). Instead, as shown in Fig.~\ref{results}, by adjusting the prior probabilities based on concept exposure the dynamic model is able to capture learning time patterns in the target groups ($R^{2}=0.96$ compared to $R^{2}=0.73$ for the static model). Expectedly, both models perform similarly in the control groups as they were designed to not encourage the use of any particular operator ($R^{2}=0.72$; $R^{2}=0.71$ for the static model). The impact of the learning capability of the model is most evident in the target group concept sequence, which was designed to this effect. If the structure of the concepts does not bias the LoT primitives one way or the other, it is expected that a static model will provide a reasonable fit. However, it is difficult to tell a priori how unbiased a set of concepts really is, so experiments relying on repeated concept exposure should always take between-concept learning into account.

\begin{figure}
        \centering
        \includegraphics[scale=1]{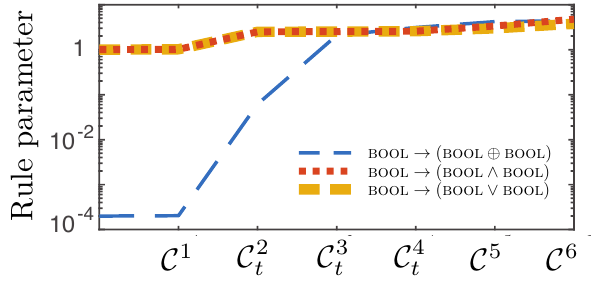}
        \caption{Evolution of Dirichlet parameters of different rules after each concept experienced by the target group.}
       \label{evol}
\end{figure}

Allowing the model to constantly update its beliefs from concept to concept is a requisite to capture human learning times. We now explain how the pattern of subjective difficulties in Fig.~\ref{results} emerged in the \textit{Dynamic model}. In this scenario, learning for the model is formalized by the update of rule parameters from concept $t$ to concept $t+1$ according to \eqref{Dirichlet}. In Fig.~\ref{evol} we show how this learning takes place in the concept sequence for the target group. There are mainly two competing formulas when \targetb is presented: $x_{i} \oxor x_{j}$ and $(x_{i} \oand \neg x_{j}) \oor (\neg x_{i} \oand x_{j})$. Given the low a priori value of the parameter of the $\oxor$ rule, the posterior of the formulas of type $(x_{i} \oand \neg x_{j}) \oor (\neg x_{i} \oand x_{j})$, which do not use the $\oxor$ operator, is much higher than the posterior of $x_{i} \oxor x_{j}$. Therefore, in Fig.~\ref{results} (middle top) we see a large predicted difficulty by the dynamic model for this concept (since the posterior lies mainly over these longer formulas without $\oxor$, see  \eqref{expected length}). 

However, the little increment in the $\oxor$ rule after \targetb (see Fig.~\ref{evol}) is sufficient for making the formula $x_{k} \oxor x_{l}$ to have higher relative posterior in the next concepts, making the increment in the parameter of the $\oxor$ rule much greater than before. Additionally, the difficulty inferred by the model is much smaller the second time the concept is presented (compare \targetd and \targetb concepts in Fig.~\ref{results}), since now the posterior is more evenly distributed between long (without $\oxor$) and short (with $\oxor$) formulas (see \eqref{expected length}). Finally, when the concept \testa is presented, the learner has completely compiled the $\oxor$ rule into her language, ascribing the formulas that use the $\oxor$ operator a much higher posterior probability relative to the long formulas that do not use the $\oxor$ operator. Therefore, the inferred difficulty for \testa is much smaller than those describing previous concepts, almost as simple as concept \targeta (see Fig.~\ref{results}).

\section{Discussion}

\par We measured the subjective difficulty that participants experience when learning a sequence of concepts. To explain this subjective difficulty, we resource to propositional logic as a  base description language. In the target group we experimented with concepts which can be succinctly described in the base language {\em that also contains an extra operator $\oxor$} for exclusive disjunction but that needed necessarily longer descriptions over the base language (where this operator is absent). On the contrary, the control group is exposed to concepts where $\oxor$ does not help to achieve succinctness.

Learning times are consistent with the hypothesis that participants in the target group smoothly adopt the $\oxor$ as a new primitive of their LoT in order to absorb the concepts they have been exposed to, with no more incentive than decreasing the expected complexity of future concepts. We do not claim that participants have learned the $\oxor$ operator defined by any specific formula using the previous operators, however, their LoT seems to have constructed an operation that matches the semantics of the exclusive or in order to compress such patterns of data and identify them more efficiently.

\par Here, we focus on transfer learning effects when learning sequential concepts that share the same hierarchical structure. We acknowledge, however, that several other transfer learning effects are present in human sequential logical concept learning, such as when subsequent concepts differ in the relevant variables (e.g.\  color lights in our experiment) \cite{blair2009extremely}, when changing the relevant variables in subsequent exclusive disjunctions \cite{kruschke1996dimensional}, or when two categories are learned in an interleaved or a focused manner \cite{carvalho2014putting}. However, unlike superficial knowledge about the task (like the frequency of appearance of different symbols and logical operators in the concept sequence), identifying the latent hierarchical structure of concepts have extremely important computational consequences: it allows for exponentially less complex representations \cite{bengio2013representation,lake2015human}, maximizing the expected value of future computations within resource-bounded constraints~\cite{gershman2015computational}. In our task, in order to focus primarily on the learning process of the $\oxor$ structure, we randomize variables in each trial, such that other kinds of transitions are averaged out across participants. 

\par Most LoT studies provide a language that is fixed once trained or inferred over a specific data. We claim that when a specific language beats a second one at fitting some experimental data, what we may be seeing is an effect of prior experience (including from the experiment itself), more than an intrinsic feature of the LoT. This leads to a fundamental difficulty in trying to experimentally uncover what the actual human symbolic substrate of thought is. Experimental results have shown for instance that a grammar with \textit{and, or}, and \textit{not} better explains Boolean concept learning than one with \textit{nand}, despite both being expressively equivalent~\cite{piantadosi2016logical}.  In our view, this cannot be taken to mean anything more than that in the current state of affairs of the world, the \textit{nand} operator is not very useful for compressing information. We have shown that participants can rapidly compile new expressions in their LoT if they begin to be useful, which emphasizes that one cannot simply ignore the order in which concepts are presented to the participant when studying aspects of the LoT.

When Fodor proposed the Language of Thought hypothesis \cite{fodor1975language}, what he had in mind was a symbolic system we all came equipped with from birth. Stating that this language is in fact always flexible might seem in outright contradiction with Fodor's original idea. In fact, what studies in the LoT literature (including this one) are probably probing is one among many languages in a hierarchy of increasing abstraction. As we progress in life, we find some conceptual summaries useful, and compiled them in a more abstract token. It is even likely that there is no proper hierarchy with sharply defined boundaries between levels, but instead a less organized progression of concepts of increasing abstraction, with thought progressing seamlessly using constructs at different levels. 

\section{Conclusion}

We defined a model to measure the subjective difficulty of learning a sequence of concepts. The model updates the grammar production probabilities between concepts and predicts difficulty as the size of compatible formulas weighted by their posterior probability. This learning mechanism allows to simulate the emergence of a new primitive in the language, as it becomes useful to encode the concepts presented so far. The predicted difficulties strongly resembles the pattern of human learning times in a sequence of concepts that required the $\oxor$ operator in order to be efficiently represented.
\bibliography{pre}
\end{document}